\documentclass[runningheads]{llncs}
\usepackage{graphicx}

\usepackage{tikz}
\usepackage{comment}
\usepackage{amsmath,amssymb} 
\usepackage{color}

\usepackage{graphicx}
\usepackage{amsmath}
\usepackage{amssymb}
\usepackage{booktabs}

\usepackage{multirow}
\usepackage{tabularx,verbatim}
\usepackage{xspace}
\usepackage{algorithm}
\usepackage{algorithmic}
\usepackage{setspace}
\usepackage{comment}
\usepackage{bbm}
\usepackage{enumitem}
\usepackage{colortbl}
\usepackage{color, xcolor} 
\usepackage[T1]{fontenc}
\usepackage{bbm}
\usepackage{listings}

\usepackage{pifont}
\newcommand{\cmark}{\ding{51}}%
\newcommand{\xmark}{\ding{55}}%

\def\eg{\emph{e.g.}} 
\def\ie{\emph{i.e.}}

\definecolor{remark}{rgb}{1,.5,0} 
\definecolor{citecolor}{rgb}{0,0.443,0.737} 
\definecolor{linkcolor}{rgb}{0.956,0.298,0.235} 
\definecolor{gray}{gray}{0.5}
\definecolor{cyan}{rgb}{0.831,0.901,0.945}
\definecolor{teal}{rgb}{0,0.5,0.5}
\definecolor{lightskyblue}{rgb}{0.53,0.8,0.976}

\usepackage{xspace}
\newcommand{\ours}{SHADE\xspace}

\usepackage{amsmath}

\usepackage[pagebackref,breaklinks,colorlinks,bookmarks=false]{hyperref}
\colorlet{dark-blue}{blue!70!black}
\colorlet{dark-green}{green!60!black}
\colorlet{dark-red}{red!80!black}
\definecolor{mypink}{RGB}{219, 48, 122}
\hypersetup{
 colorlinks=true,%
 citecolor=citecolor,%
 filecolor=citecolor,%
 linkcolor=linkcolor,%
 urlcolor=magenta
}
\usepackage{url}

\usepackage[capitalize]{cleveref}
\crefname{section}{Sec.}{Secs.}
\Crefname{section}{Section}{Sections}
\Crefname{table}{Table}{Tables}
\crefname{table}{Tab.}{Tabs.}

\usepackage[accsupp]{axessibility}  

\begin{document}
\pagestyle{headings}
\mainmatter
\def\ECCVSubNumber{86}  

\title{Style-Hallucinated Dual Consistency Learning for Domain Generalized Semantic Segmentation}

\titlerunning{SHADE}
%
\author{Yuyang Zhao\inst{1} \and Zhun Zhong\inst{2} \and Na Zhao\inst{1} \and Nicu Sebe\inst{2} \and Gim Hee Lee\inst{1}\index{Lee, Gim Hee}}
\authorrunning{Y. Zhao et al.}
\institute{National University of Singapore \and
University of Trento \\
\url{https://github.com/HeliosZhao/SHADE}}
\maketitle

\begin{abstract}

In this paper, we study the task of synthetic-to-real domain generalized semantic segmentation, which aims to learn a model that is robust to unseen real-world scenes using only synthetic data. The large domain shift between synthetic and real-world data, including the limited source environmental variations and the large distribution gap between synthetic and real-world data, significantly hinders the model performance on unseen real-world scenes. In this work, we propose the \textbf{S}tyle-\textbf{HA}llucinated \textbf{D}ual consist\textbf{E}ncy learning (\textbf{\ours}) framework to handle such domain shift. Specifically, \ours is constructed based on two consistency constraints, Style Consistency (SC) and Retrospection Consistency (RC). SC enriches the source situations and encourages the model to learn consistent representation across style-diversified samples. 
RC leverages real-world knowledge to prevent the model from overfitting to synthetic data and thus largely keeps the representation consistent between the synthetic and real-world models. 
Furthermore, we present a novel style hallucination module (SHM) to generate style-diversified samples that are essential to consistency learning. 
SHM selects basis styles from the source distribution, enabling the model to dynamically generate diverse and realistic samples during training. 
Experiments show that our \ours yields significant improvement and outperforms state-of-the-art methods by 5.05\% and 8.35\% on the average mIoU of three real-world datasets on single- and multi-source settings, respectively.

\end{abstract}

\section{Introduction}
\begin{sloppypar}
Semantic segmentation, \ie, predicting a semantic category for each pixel, plays a crucial role in autonomous driving.
With modern deep neural networks~\cite{liu2021swin,liu2022convnet} and abundant annotated data~\cite{cityscapes,imagenet}, fully-supervised methods~\cite{deeplab,long2015fully,xie2021segformer} have achieved remarkable success on many public datasets. Nonetheless, annotating each pixel for a high-resolution image is expensive and time-consuming. For example, it takes more than 1.5 hours to annotate a 1024$\times$2048 driving scene~\cite{cityscapes}, and the time is even doubled for scenes under adverse weather~\cite{acdc} and poor lighting conditions~\cite{darkzurich}. 
To alleviate the heavy annotation cost, unsupervised domain adaptive semantic segmentation (UDA-Seg)~\cite{adaptseg,zhou2021context,zhou2021domain} has been introduced to learn models that can perform well on a target set, given the labeled (source) data and unlabeled (target) data. 
Considering that synthetic data can be automatically generated and annotated by a pre-designed engine~\cite{gtav,synthia}, existing works of UDA-Seg commonly use synthetic data as the source domain and real-world data as the target domain.
\end{sloppypar}

\begin{figure}[t]
    \centering
    \includegraphics[width=.9\textwidth]{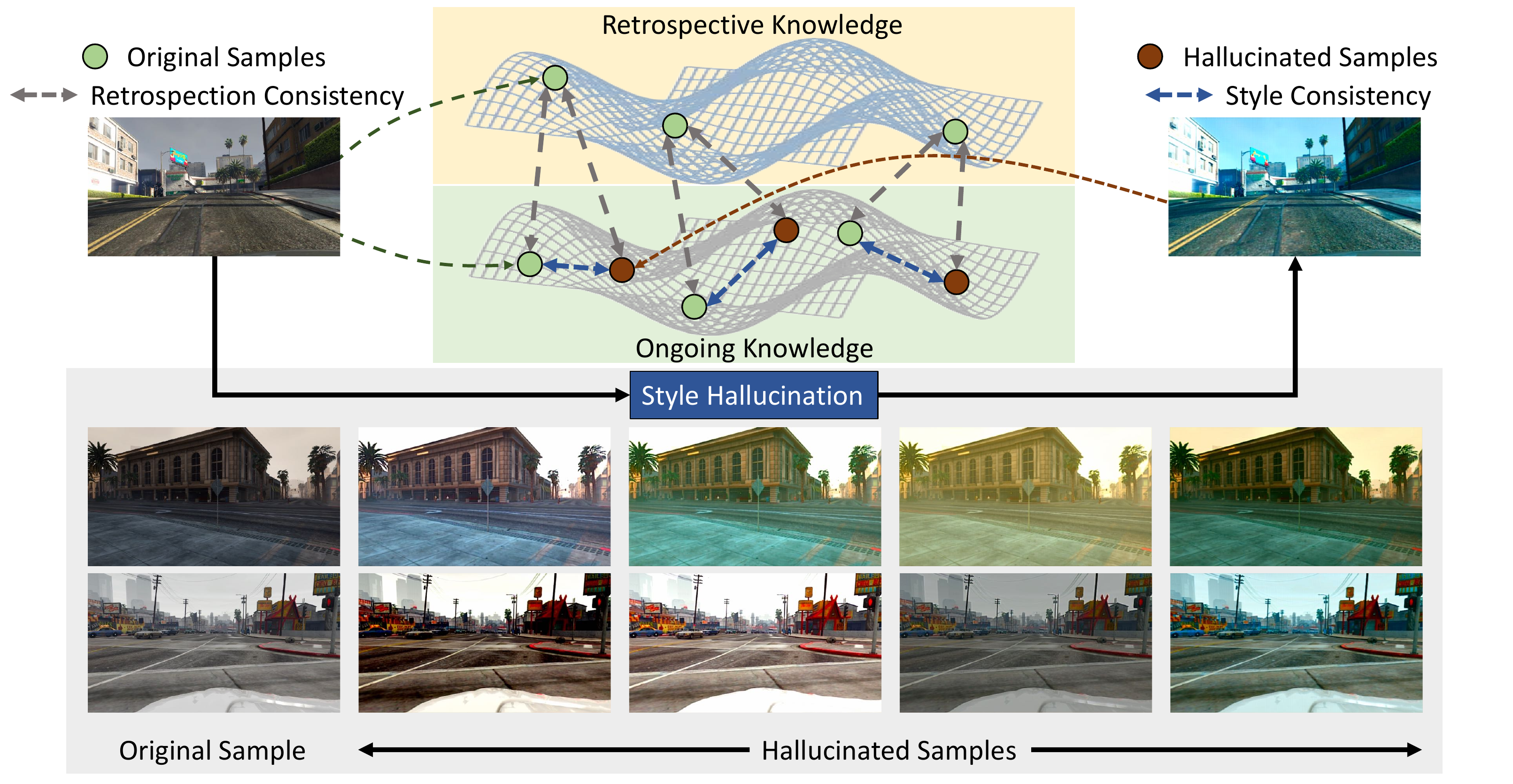}
    \caption{Illustration of dual consistency constraints and examples of style hallucination. We generate hallucinated samples (\textcolor{brown}{brown circle}) from the style hallucination module and then utilize the paired samples and real-world (retrospective) knowledge to learn style consistency (\textcolor{dark-blue}{blue dash line}) and retrospection consistency (\textcolor{gray}{gray dash line}).}
    \label{fig:intro}
\end{figure}

Despite the success of UDA-Seg, the limitations are still severe. First, the adapted model achieves remarkable performance on the target data, but  
still degrades when facing unseen out-of-distribution samples. Second, it is almost impractical to collect diverse enough real-world data that can cover all the conditions. 
For example, a model trained with Singapore scenes cannot well address the snowy road in Switzerland. 
To mitigate such limitations, a more practical but challenging setting, domain generalized semantic segmentation (DG-Seg)~\cite{robustnet,DRPC,zhong2022adversarial}, is introduced in the community. 
DG-Seg only leverages annotated source data to train a robust model that can cope with different unseen conditions. Similar to UDA-Seg, synthetic data are commonly used as the source data to release expensive annotation cost in DG-Seg. In this paper, we aim to solve the synthetic-to-real DG task in semantic segmentation.

The main challenge for synthetic-to-real DG-Seg is to cope with the significant domain shift between source and unseen target domains, which can be roughly divided into two aspects. First, the environmental variations in source data are very limited compared to those of unseen target data. 
Second, there exists large distribution gap between synthetic and real-world data, \textit{e.g.,} image styles and characteristics of objects.
To learn the domain-invariant model that can address the domain shift, 
previous works design tailor-made modules~\cite{robustnet,ibn} to remove domain-specific information, or leverage extra real-world data to transfer synthetic data~\cite{FSDR,DRPC} to real-world styles for narrowing the distribution gap.
However, the removal of domain-specific information~\cite{robustnet,ibn} is not complete and explicit due to the lack of real-world information; the real-world style transfer~\cite{FSDR,DRPC} heavily relies on extra data, which are not always available in practice, and ignores the invariant representation within the source domain.
Taking the above into account, in this paper, we aim to explicitly learn domain-invariant representation with the stylized samples in the source domain and bridge the gap between synthetic and real-world data without using extra real-world data.

To this end, we introduce a novel dual consistency learning framework that can jointly achieve the above two goals.
As shown in Fig.~\ref{fig:intro}, we introduce two consistency constraints, \textit{style consistency} (SC) and \textit{retrospection consistency} (RC), which explicitly encourage the model to learn style invariant representation and to be less overfitting to the synthetic data respectively. Specifically, we first diversify the samples in SC by style variation, and then enforce the consistency between them.
To obtain the diverse samples, we adopt the style features, \ie, channel-wise mean and standard deviation, to generate new data.
Compared with directly transferring the whole image (\eg, CycleGAN~\cite{cyclegan}), changing style features can maintain the pixel alignment to the utmost extent. This forces the model to learn pixel-level consistency between paired samples.
In addition, our RC lies in the guidance of real-world information. 
We leverage ImageNet~\cite{imagenet} pre-trained model which is available in all the DG-Seg models.
The pre-trained model contains general knowledge of classifying some ``things'' class, \eg, bicycle, bus and car. Furthermore, the features of these classes in the pre-trained model reflect
the representation in the context of real-world.
Consequently, such features can serve as the guidance for the ongoing model to retrospect 
what the real-world looks like
and to lead the model less overfitting to the synthetic data.

Style diversifying is crucial for the success of dual consistency learning.
Previous works~\cite{crossnorm,zhou2021mixstyle} commonly mix or swap styles within the source domain, which will generate more samples of the dominant styles (\eg, daytime in GTAV~\cite{gtav}). Nevertheless, it is not the best way since the target styles may be quite different from the dominant styles. 
To fully take the advantage of all the source styles, we propose style hallucination module (SHM), which leverages $C$ basis styles to represent the style space of $C$ dimension and thus to generate new styles. 
Ideally, the basis styles should be linearly independent so the linear combination of basis styles can represent all the source styles. 
However, many unrealistic styles that impair the model training are generated when we directly take $C$ orthogonal unit vectors as the basis. To reconcile diversity and realism, we use farthest point sampling~(FPS)~\cite{qi2017pointnet++} to select $C$ styles from all the source styles as basis styles. 
Such basis styles contain many rare styles since rare styles are commonly far away from the dominant ones. With these basis styles that represent the style space in a better way, we utilize linear combination to generate new styles.
To summarize, we propose the \textbf{S}tyle-\textbf{HA}llucinated \textbf{D}ual consist\textbf{E}ncy learning (\textbf{\ours}) framework for domain generalization in the context of semantic segmentation, and our contributions are as follows:
\begin{itemize}
    \item We propose the dual consistency constraints for domain generalized semantic segmentation, which learn the style invariant representation among diversified styles and narrow the domain gap between synthetic and real domains by retrospecting the real-world knowledge.
    \item We propose the style hallucination module to generate new and diverse styles with the representative basis styles, enhancing the dual consistency learning.
    \item Experiments on single-source and multi-source DG benchmark demonstrate the effectiveness of the proposed method and we achieve new state-of-the-art performance.
\end{itemize}

\section{Related Work}

\noindent\textbf{Domain Generalization.} 
To tackle the performance degradation in the out-of-distribution conditions, domain generalization~\cite{FSDR,DRPC,robustnet,meng2022attention,zhou2021mixstyle} is introduced to learn a robust model with one or multiple source domains, aiming to perform well on unseen domains. Considering the expensive annotation cost in semantic segmentation, synthetic data are commonly adopted as the source domain in recent DG-Seg works. 
The large domain shift and the restricting conditions in the training data greatly limit the performance on the unseen real-world test data.
To address these problems, one main stream of previous works~\cite{FSDR,DRPC} focuses on diversifying training data with real-world templates and learning the invariant representation from all the domains. Another main stream aims at directly learning the explicit domain-invariant features within the source domain~\cite{robustnet,ibn,crossnorm}. IBN-Net~\cite{ibn} and ISW~\cite{robustnet} leverage tailor-made instance normalization block and whitening transformation to reduce the impact of domain-specific information. 
Different from previous methods, we generate new styles only with the synthetic source domain and learn the invariant representation across styles.

\noindent\textbf{Consistency learning (CL).}
CL is adopted by many computer vision tasks and settings. One main stream is leveraging CL to exploit the unlabeled samples in semi-supervised learning~\cite{french2019semi,tarvainen2017mean} and unsupervised domain adaptation~\cite{pan2020unsupervised,zou2018unsupervised,zhou2022uncertainty} since the consistency between two view samples can be used for pseudo-labeling. CL is also applied to address the corruptions and perturbations~\cite{hendrycks2020augmix,wang2021augmax} by maximizing the similarity between different augmentations.
In addition, CL is also used in self-supervised learning~\cite{simclr,moco} as the contrastive loss to utilize totally unlabeled data.
We introduce CL into domain generalization, leading the model robust to various styles. We also leverage consistency with real-world knowledge to narrow the domain gap between synthetic and real-world data. 

\noindent\textbf{Style Variation.}
Style features are widely explored in style transfer~\cite{dumoulin2016learned,adain}, which aims at changing the image style but maintaining the content. 
Inspired by this, recent domain generalization methods leverage the style features to generate diverse data of different styles to improve the generalization ability.
Swapping~\cite{crossnorm,zhao2022sfocda} and mixing~\cite{zhou2021mixstyle} existing styles within the source domains is an effective way and generating new styles~\cite{wang2021L2D} by specially designed modules can also make sense. 
We also only leverage the styles within the source domain but take the relatively rare styles in the source domains into account, thus generating more diverse samples to improve generalization ability.

\begin{figure}[t]
    \centering
    \includegraphics[width=.9\textwidth]{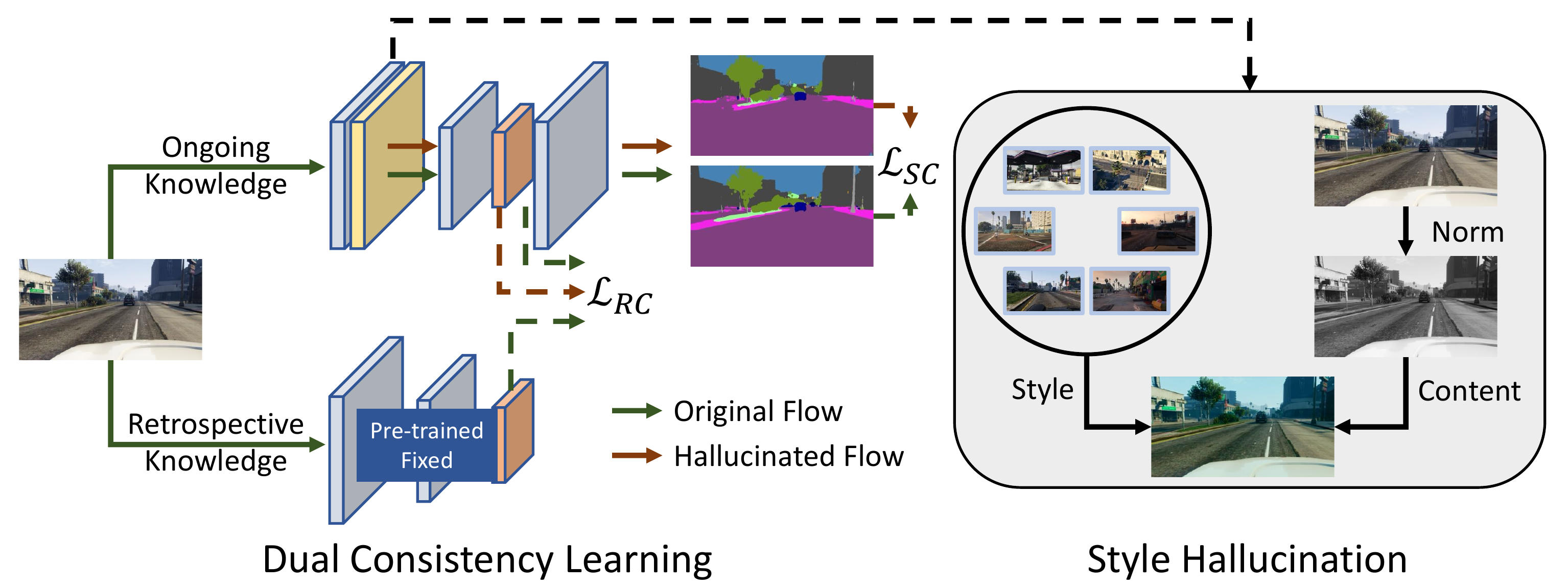}
    \caption{Overview of the proposed style-hallucinated dual consistency learning (\ours) framework. Training images are fed into the segmentation model (ongoing knowledge) and the ImageNet pre-trained model (retrospective knowledge). 
    The style hallucination module is inserted into a certain layer of the segmentation model to generate stylized samples.
    Finally, the model is optimized by the dual consistency losses: style consistency $\mathcal{L}_{SC}$ and retrospection consistency $\mathcal{L}_{RC}$. Note that the cross entropy loss is also used and we omit it here for brevity.}
    \label{fig:framework}
\end{figure}

\section{Methodology}
\noindent\textbf{Preliminary.} In synthetic-to-real domain generalized semantic segmentation, one or multiple labeled source domains $\mathcal{S}=\{x_S^i, y_S^i\}_{i=i}^{N_S}$, where ${N_S}$ is the number of source domains, are used to train a segmentation model, which is deployed to unseen real-world target domains $\mathcal{T}$ directly. 
In general, the source and target domains share the same label space $Y_S, Y_T \in (1, N_C)$ but belong to different distributions. The goal of this task is to improve the generalization ability of model in unseen domains using only the source data.

\begin{sloppypar}
\noindent\textbf{Overview.} 
To solve the above challenging problem, we propose the \textbf{S}tyle-\textbf{HA}llucinated \textbf{D}ual consist\textbf{E}ncy learning (\textbf{\ours}) framework, which is quipped with dual consistency constraints and a Style Hallucination Module (SHM). The dual consistency constraints effectively learn the domain-invariant representation and reduce the gap between the real and synthetic data. SHM enriches the training samples by dynamically generating diverse styles, which catalyzes the advantage of dual consistency learning. The overall framework is shown in Fig.~\ref{fig:framework}.
\end{sloppypar}

\subsection{Dual Consistency Constraints}
In \ours, we introduce two consistency learning constraints: (1) Style Consistency (SC) that aims at learning the consistent predictions across stylized samples. (2) Retrospection Consistency (RC) that focuses on narrowing the distribution shift between synthetic and real-world data in the feature-level.

\noindent\textbf{Style Consistency (SC).}
Different from traditional cross entropy constraint focusing on the high-level semantic information, logit pairing~\cite{hendrycks2020augmix,kannan2018adversarial} has demonstrated its effectiveness in learning adversarial samples by highlighting the most invariant information.
Inspired by this, we propose SC to ameliorate the style shift with logit pairing.
To fulfill this, SC requires the style-diversified samples $\Tilde{x}_S$ which share the same semantic content with the source samples $x_S$ but of different styles. Due to the pixel-level segmentation requirement, it is better to obtain the style-diversified samples $\Tilde{x}_S$ by non-geometry style variation, \eg, MixStyle~\cite{zhou2021mixstyle}, CrossNorm~\cite{crossnorm} and the proposed style hallucination in Sec.~\ref{sec:sh}.
Formally, we minimize the Jensen-Shannon Divergence (JSD) between the posterior probability $p$ 
of the semantically aligned $\Tilde{x}_S$ and $x_S$:
\begin{equation}
\begin{aligned}
    \mathcal{L}_{SC} (x_S, \Tilde{x}_S) &= JSD\left(p(x_S); p(\Tilde{x}_S)\right) \\
    &= \frac{1}{2} \left(D [p(x_S) || Q] + D [p(\Tilde{x}_S) || Q] \right), \\
\end{aligned}
\end{equation}
where $Q=(p(x_S) + p(\Tilde{x}_S))/2$ is the average information of the original and style-diversified samples. $D$ denotes the KL Divergence between the posterior probability $p \in \{p(x_S), p(\Tilde{x}_S)\}$ and $Q$. 
JSD highlights the invariant pixel-level semantic information across two styles, impelling the model to be stable and insensitive to varied styles.

\noindent\textbf{Retrospection Consistency (RC).}
Backbones of semantic segmentation models commonly start from ImageNet~\cite{imagenet} pre-trained weights since the pre-trained backbones have learned general representation of many ``things'' categories in the real-world, including some classes in the self-driving scenes, \eg, bicycle, bus, and car. However, the model 
learns more task-specific information and fit to the synthetic data when training with synthetic semantic segmentation data.
As the model is required to deploy on unseen real-world scenarios, it is important to obtain some knowledge for the real-world objects.
Previous works~\cite{DRPC,FSDR} leverage the abundant and even carefully selected images from real-world data, which may not be readily fulfilled in applications. 
Since ImageNet pre-trained weights are available to every segmentation model as the initialization, we propose RC to leverage such knowledge to lead the model less overfitting to the synthetic data and to retrospect the real-world knowledge lying in initialization.
As the ImageNet pre-trained model is trained by image classification, only ``things'' classes are learned instead of ``stuff'' classes like road and wall. Consequently, RC is implemented as the feature-level distance minimization on those pixels of ``things'' classes.
In addition, the style-diversified samples $\Tilde{x}_S$ in style consistency is also used in RC, which can lead the generated samples close to the real-world style.
We define RC as:

\begin{equation}
\small
\begin{aligned}
    \mathcal{L}_{RC}(x_S, \Tilde{x}_S) = \frac{1}{\sum_m M_{things}^{(m)}} \sum_m M_{things}^{(m)} \cdot & \left( \left( f(x_S;\theta_S)^{(m)} - f(x_S;\theta_{IN})^{(m)} \right)^2 \right. \\ 
    +& \left. \left( f(\Tilde{x}_S;\theta_S)^{(m)}-f(x_S;\theta_{IN})^{(m)} \right)^2 \right),
\end{aligned}
\end{equation}
where $M_{things}$ denotes the mask of ``things'' classes. $f(x_S;\theta_S)$ and $f(\Tilde{x}_S;\theta_S)$ denote the bottleneck feature of original sample $x_S$ and style-diversified sample $\Tilde{x}_S$ respectively, which are obtained from the ongoing segmentation model $\theta_S$. $f(x_S;\theta_{IN})$ denotes the bottleneck feature of original sample $x_S$ in the retrospective ImageNet pre-trained model $\theta_{IN}$.

\noindent\textbf{Discussion.} Our retrospection consistency is inspired by the Feature Distance (FD) in DAFormer~\cite{hoyer2021daformer} but with different motivation and implementation. \textit{First}, DAFormer focuses on unsupervised domain adaptation in semantic segmentation with unlabeled real-world data available, so it only focuses on fitting to the specific target domain (CityScapes~\cite{cityscapes}) rather than addressing unseen domain shift. \textit{Second}, FD in DAFormer is used to better classify those similar classes (\eg, bus and train) with the classification knowledge from ImageNet. %
Since we have no idea about the target distribution in DG-Seg, RC in our framework serves as an important guidance for the real-world knowledge, especially for more complex real-world scenes (\textit{e.g.}, BDD100K~\cite{bdd} and Mapillary~\cite{mapillary}). 
In addition, RC can also enhance the learning of real-world styles by taking the style-diversified samples into account.

\subsection{Style Hallucination}
\label{sec:sh}
\noindent\textbf{Background.} Style transfer~\cite{chen2021diverse,adain} and domain generalization~\cite{crossnorm,zhou2021mixstyle} methods show that the channel-wise mean and standard deviation can represent the non-content style of the image, which plays an important role in the domain shift. The style features can be readily used by AdaIN~\cite{adain} which can transfer the image to an arbitrary style while remaining the content:
\begin{equation}
    \text{AdaIN}(x, y)=\sigma(y)\left(\frac{x-\mu(x)}{\sigma(x)}\right)+\mu(y),
\end{equation}
where $x$ and $y$ denotes the feature maps providing the content and style respectively. $\mu(*)$ and $\sigma(*)$ denotes the channel-wise mean and standard deviation. In domain generalization, as only one or multiple source domains are accessible, previous works modify AdaIN by replacing style features with other source styles. Those styles can be directly obtained from other samples~\cite{crossnorm} or can be generated by mixing other styles with its own styles~\cite{zhou2021mixstyle}.

\noindent\textbf{Style Hallucination Module (SHM).}
Unlike image classification benchmarks where images are commonly of the same style in one dataset, even one semantic segmentation dataset contains various styles, \eg, daytime, nighttime and twilight.
This is why existing style variation methods~\cite{crossnorm,zhou2021mixstyle} can work in single-source domain generalized segmentation.
However, the ways of these methods in generating extra styles are sub-optimal, since they just randomly swap or mix source styles without considering the frequency and diversity of the source styles. As a result, more samples of the dominant style (\eg, daytime) will be generated, yet the generated distribution may be quite different from the real-world one. 
Since we have no idea about the real-world target set, it is better to diversify the source samples as much as possible. We next introduce the Style Hallucination Module (SHM) for generating diverse source samples.

\noindent\textbf{Definition 1:} 
\textit{A basis $B$ of a vector space $V$ over a field $F$ is a linearly independent subset of $V$ that spans $V$. When the field is the reals $\mathbb{R}$, the resulting basis vectors are $n$-tuples of reals that span $n$-dimensional Euclidean space $\mathbb{R}^n$}~\cite{basis}.

According to Definition 1, style space can be viewed as a subspace of $C$-dimensional vector space, and thus all possible styles can be represented by the basis vectors.
However, if we directly take $C$ linearly independent vectors as the basis, \eg, orthogonal unit vectors, many unrealistic styles %
are generated since the realistic styles are only in a small subspace, and such generated styles %
can impair the model training. 
To reconcile the diversity and realism, we use farthest point sampling~(FPS)~\cite{qi2017pointnet++} to select $C$ styles from all the source styles as basis styles. 
FPS is widely used for point cloud downsampling, which can iteratively choose $C$ points from all the points, such that the chosen points are the most distant points with respect to the remaining points.
Despite not strictly linearly independent, basis styles obtained by FPS can represent the style space to the utmost extent, and also contain many rare styles since rare styles are commonly far away from dominant ones. 
In addition, we recalculate the dynamic basis styles every $k$ epochs instead of fixing basis style, as the style space is changing along with the model training. To generate new styles, we sample the combination weight $W=[w_1, \cdots, w_C]$ from Dirichlet distribution $B([\alpha_1, \cdots, \alpha_C])$ with the concentration parameters $[\alpha_1, \cdots, \alpha_C]$ all set to $1/C$. 
The basis styles are then linearly combined by $W$:
\begin{equation}
    \mu_{HS} = W\cdot \mu_{base}, \qquad \sigma_{HS} = W \cdot \sigma_{base},
\end{equation}
where $\mu_{base}\in \mathbb{R}^{C\times C}$ and $\sigma_{base} \in \mathbb{R}^{C\times C}$ are the $C$ basis styles. With the generated styles, style hallucinated samples $\Tilde{x}_S$ can be obtained by:
\begin{equation}
    \Tilde{x}_S = \sigma_{HS}\left(\frac{x_S-\mu(x_S)}{\sigma(x_S)}\right)+\mu_{HS}.
\end{equation}

\begin{figure}[t]
    \centering
    \includegraphics[width=.9\textwidth]{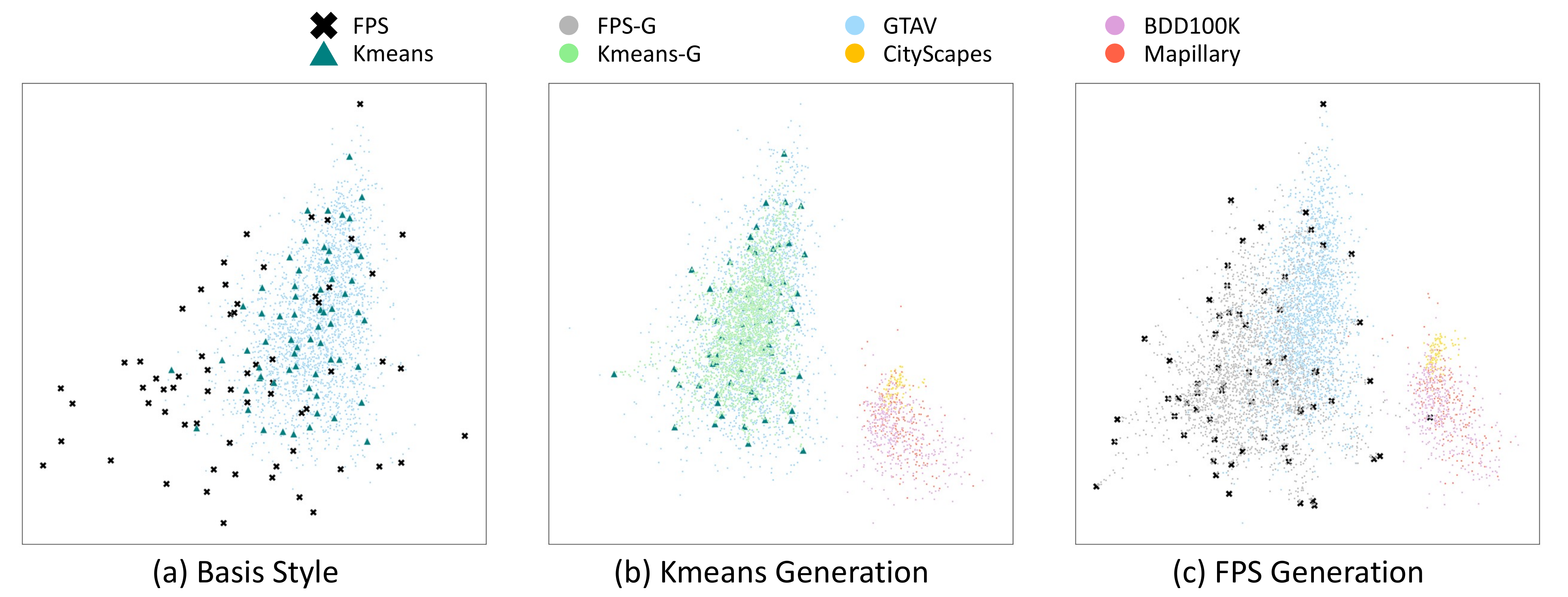}
    \caption{Visualization of distributions of different domains. (a) Comparison of two kinds of basis styles; (b) generated style with Kmeans basis style; (c) generated style with FPS basis style. (Zoom in for details.) }
    \label{fig:visual-dist}
\end{figure}

\noindent\textbf{Discussion.}
Selecting representative basis styles is crucial for SHM. FPS is adopted in our method as it can cover the rare styles to the utmost extent. Another way is taking the Kmeans~\cite{macqueen1967some_kmeans} clustering centers as the basis. As shown in Fig.~\ref{fig:visual-dist}(a), FPS samples (\textcolor{black}{black cross}) spread out more than Kmeans centers (\textcolor{teal}{teal triangle}), and can cover almost all possible source styles (\textcolor{lightskyblue}{lightskyblue point}). When using the basis styles for style generation, styles obtained from Kmeans centers (Fig.~\ref{fig:visual-dist}(b)) are still within the source distributions and even ignore some possible rare styles. In contrast, FPS basis styles can generate more diverse styles (Fig.~\ref{fig:visual-dist}(c)), and even generate some styles close to the real-world ones (\textcolor{yellow}{yellow}, \textcolor{pink}{pink} and \textcolor{orange}{orange} point). Tab.~\ref{table:style-variation} further demonstrates the effectiveness of FPS basis styles and shows that the Kmeans basis styles are even worse than directly swapping and mixing source styles.

\subsection{Training Objective}
The overall training objective is the combination of pixel-level cross entropy loss and the proposed two consistency constraints:
\begin{equation}
\begin{aligned}
    \mathcal{L} &= \frac{1}{2} \left( \mathcal{L}_{CE}(x_S,y_S) + \mathcal{L}_{CE}(\Tilde{x}_S,y_S) \right) \\
    &+ \lambda_{SC} \mathcal{L}_{SC}(x_S, \Tilde{x}_S) + \lambda_{RC} \mathcal{L}_{RC}(x_S, \Tilde{x}_S),
\end{aligned}
\end{equation}
where $\lambda_{SC}$ and $\lambda_{RC}$ are the weights for style consistency and retrospection consistency, respectively.

\section{Experiments}

\subsection{Experimental Setup}
\label{sec:seg_imple_details}

\noindent\textbf{Datasets.}
Two synthetic datasets (GTAV~\cite{gtav} and SYNTHIA~\cite{synthia}) are taken as the source domains.
GTAV~\cite{gtav} contains 24,966 images with the size of 1914$\times$1052, splitting into 12,403 training, 6,382 validation, and 6,181 testing images. SYNTHIA~\cite{synthia} contains 9,400 images of 960$\times$720, where 6,580 images are used for training and 2,820 images for validation.
We evaluate the model on the validation sets of the three real-world datasets. CityScapes~\cite{cityscapes} contains 500 validation images of 2048$\times$1024. BDD-100K~\cite{bdd} and Mapillary~\cite{mapillary} contain 1,000 1280$\times$720 images and 2,000 1920$\times$1080 images for validation, respectively.

\noindent\textbf{Implementation Details.} Following \cite{robustnet}, we use DeepLabV3+~\cite{deeplab} as the segmentation model. The segmentation model is equipped with two backbones, ResNet-50 and ResNet-101~\cite{he2016deep}. The SHM is inserted after the first Conv-BN-ReLU layer (layer0). We re-select the basis styles with the interval $k=3$. We set $\lambda_{SC}=10$ and $\lambda_{RC}=1$.
Models are optimized by the SGD optimizer with the learning rate 0.01, momentum 0.9 and weight decay 5$\times$10$^{-4}$. The polynomial decay~\cite{liu2015parsenet} with the power of 0.9 is used as the learning rate scheduler. All models are trained with the batch size of 8 for 40K iterations. 
During training, four widely used data augmentation techniques are adopted, including color jittering, Gaussian blur, random flipping, and random cropping of 768$\times$768. 

\noindent\textbf{Protocols.} We conduct experiments on both single-source DG and multi-source DG. For \textit{single-source DG}, to conduct a fair comparison with \cite{robustnet} and \cite{FSDR}, we train the model with GTAV training data (12,403 images) when using ResNet-50 backbone (same with \cite{robustnet}), and with the whole GTAV datasets (24,966 images) when using ResNet-101 backbone (same with \cite{FSDR}). For \textit{multi-source DG}, we follow \cite{robustnet} to train the model with the training set of GTAV (12,403 images) and SYNTHIA (6,580 images) using ResNet-50 backbone.
\textbf{Note that} several state-of-the-art works, \textit{e.g.,} \cite{FSDR} and \cite{DRPC}, select the \textit{best checkpoint} for each target dataset respectively, which is impractical since we cannot estimate the model performance on unseen domains in real-world applications. Instead, we directly use the \textit{last checkpoint} to evaluate the three target datasets, which is more in line with the practical purpose of domain generalization.

\noindent\textbf{Baseline.} The baseline in each protocol is the model trained with the corresponding source training data by cross entropy loss function.

\noindent\textbf{Evaluation Metric.} 
We use the 19 shared semantic categories for training and evaluation. The mean intersection-over-union (mIoU) of the 19 categories on the three real-world datasets is adopted as the evaluation metric.

\begin{table}[t]
\centering
\footnotesize
\setlength{\tabcolsep}{3pt}
\caption{Comparison with state-of-the-art methods on single-source DG with ResNet-50 and ResNet-101 as backbone, respectively. 
``Extra Data'' denotes using extra real-world data during training. $^{\S}$ denotes selecting best checkpoint for each target dataset.}
\label{tab:gtav-sota}
\begin{tabular}{l|l|c|p{1.5cm}<\centering|p{1.5cm}<\centering|p{1.5cm}<\centering|p{1.5cm}<\centering}
\toprule
Net & Methods (GTAV) & Extra Data & CityScapes& BDD100K & Mapillary& Mean \\
\midrule
\multicolumn{1}{l|}{\multirow{7}{*}{\rotatebox{90}{\textbf{ResNet-50}}}} & Baseline& \xmark & 28.95 & 25.14& 28.18& 27.42\\
\cmidrule{2-7}
&SW~\cite{pan2019switchable} & \xmark &29.91 & 27.48 & 29.71 & 29.03 \\
&IterNorm~\cite{huang2019iterative} & \xmark & 31.81 & 32.70 & 33.88 & 32.79 \\
&IBN-Net~\cite{ibn} & \xmark & 33.85 & 32.30 & 37.75& 34.63\\
&DRPC~\cite{DRPC}$^{\S}$ & \textcolor{red}{\cmark} & 37.42 & 32.14& 34.12& 34.56\\
&ISW~\cite{robustnet} & \xmark & 36.58 & 35.20 & 40.33& 37.37\\
& \bf Ours & \xmark & \textbf{44.65} & \textbf{39.28} & \textbf{43.34} & \textbf{42.42} \\
\midrule
\multicolumn{1}{l|}{\multirow{6}{*}{\rotatebox{90}{\textbf{ResNet-101}}}} & Baseline & \xmark & 32.97& 30.77& 30.68& 31.47\\
\cmidrule{2-7}
& IBN-Net~\cite{ibn}  & \xmark & 37.37& 34.21& 36.81& 36.13\\
& ISW~\cite{robustnet}  & \xmark & 37.20 & 33.36& 35.57& 35.38\\
& DRPC~\cite{DRPC}$^{\S}$ & \textcolor{red}{\cmark}   & {42.53} & {38.72} & {38.05} & 39.77\\
& FSDR~\cite{FSDR}$^{\S}$ & \textcolor{red}{\cmark}  & {44.80}  & 41.20 & 43.40 & 43.13\\
& \bf Ours & \xmark & \textbf{46.66} & \textbf{43.66} & \textbf{45.50} & \textbf{45.27} \\
\bottomrule
\end{tabular}
\end{table}

\subsection{Comparison with State-of-the-art Methods}

\noindent\textbf{Single-source DG.}
In Tab.~\ref{tab:gtav-sota}, we compare \ours with state-of-the-art methods under single source setting, including SW~\cite{pan2019switchable}, IterNorm~\cite{huang2019iterative}, IBN-Net~\cite{ibn}, ISW~\cite{robustnet}, DRPC~\cite{DRPC} and FSDR~\cite{FSDR}.
\textit{First}, we compare models that are trained with GTAV training set, using ResNet-50 backbone. \ours achieves an average mIoU of 42.42\% on the three real-world target datasets, yielding an improvement of 15.00\% mIoU over the baseline and outperforming the previous best method (ISW) by 5.05\%.
\textit{Second,} we further compare methods under the training protocol of DRPC~\cite{DRPC} and FSDR~\cite{FSDR}, using ResNet-101 backbone and taking the whole set of GTAV (24,966 images) as the training data. 
Note that DRPC~\cite{DRPC} and FSDR~\cite{FSDR} utilize extra real-world data from ImageNet~\cite{imagenet} or even driving scenes. Moreover, they select the best checkpoint for each target dataset, which is impractical in the real-world applications. 
Even so, we achieve the best results on all three datasets, \textbf{46.66\% on CityScapes, 43.66\% on BDD100K, 45.50\% on Mapillary}, outperforming FSDR~\cite{FSDR} by 2.14\% in the average mIoU. These results show that we produce new state of the art in domain generalized semantic segmentation.

\noindent\textbf{Multi-source DG.} To further verify the effectiveness of \ours, we compare \ours with IBN-Net~\cite{ibn} and ISW~\cite{robustnet} under the multi-source setting.
We use ResNet-50 as the backbone and take the training set of GTAV and SYNTHIA as the source domains. As shown in Tab.~\ref{tab:multi-src}, \ours gains an improvement of 14.28\% in average mIoU over the baseline, and outperforms ISW and IBN-Net by 8.35\% and 9.84\% respectively. 
The significant improvement over ISW and IBN-Net is mainly benefited from the various samples. With richer source samples, our SHM can generate more informative and diverse styles, which can effectively facilitate the dual consistency learning.

\begin{table}[t]
\centering
\footnotesize
\setlength{\tabcolsep}{3pt}
\caption{Comparison with state-of-the-art methods on multi-source DG. All models use ResNet-50 backbone and are trained with training sets of GTAV and SYNTHIA.}
\label{tab:multi-src}
\begin{tabular}{l|p{2cm}<\centering|p{2cm}<\centering|p{2cm}<\centering|p{2cm}<\centering}
\toprule
Methods (G+S) & CityScapes& BDD100K & Mapillary& Mean \\
\midrule
Baseline & 35.46& 25.09  & 31.94  & 30.83\\
\midrule
IBN-Net~\cite{ibn}  & 35.55& 32.18  & 38.09  & 35.27\\
ISW~\cite{robustnet} & 37.69& 34.09  & 38.49  & 36.76\\
\bf Ours  & \textbf{47.43} & \textbf{40.30} & \textbf{47.60} & \textbf{45.11} \\
\bottomrule
\end{tabular}
\end{table}

\begin{table}[t]
\begin{center}
\caption{Ablation studies on loss functions. All models use ResNet-50 backbone and are trained with GTAV training set. SHM: our style hallucination module; EMA: using exponential moving average model instead of ImageNet pre-trained model.}
\label{table:ablation-loss}
\footnotesize
\setlength{\tabcolsep}{3pt}
\begin{tabular}{p{0.9cm}<\centering|p{0.9cm}<\centering|p{0.9cm}<\centering|p{0.9cm}<\centering|p{1.6cm}<\centering|p{1.6cm}<\centering|p{1.6cm}<\centering|p{1.6cm}<\centering}
\toprule
SHM & $\mathcal{L}_{SC}$ & $\mathcal{L}_{RC}$ & EMA & CityScapes & BDD100K & Mapillary & Mean \\
\midrule
\xmark & \xmark & \xmark & \xmark & 28.95 & 25.14 & 28.18 & 27.42\\
\midrule
\textcolor{dark-green}{\cmark} & \xmark & \xmark & \xmark & 38.68	& 32.40 & 35.96 & 35.68 \\
\textcolor{dark-green}{\cmark} & \textcolor{dark-green}{\cmark} & \xmark & \xmark & 42.66 & 35.92 & 40.42 & 39.67 \\
\textcolor{dark-green}{\cmark} & \xmark & \textcolor{dark-green}{\cmark} & \xmark & 41.43 & 37.65 & 41.77 & 40.29 \\
\textcolor{dark-green}{\cmark} & \textcolor{dark-green}{\cmark} & \xmark & \textcolor{dark-green}{\cmark} &42.38 & 38.04 & 42.34 & 40.92 \\
\textcolor{dark-green}{\cmark} & \textcolor{dark-green}{\cmark} & \textcolor{dark-green}{\cmark} & \xmark & \textbf{44.65} & \textbf{39.28} & \textbf{43.34} & \textbf{42.42} \\
\bottomrule
\end{tabular}
\end{center}
\end{table}

\subsection{Ablation Studies}
To investigate the effectiveness of each component in \ours, we conduct ablation studies in Tab.~\ref{table:ablation-loss}.

\noindent\textbf{Effectiveness of Style Hallucination Module (SHM).}
SHM is the basis of \ours. When using SHM only, we directly apply cross entropy loss on the style hallucinated samples. As shown in the second row of Tab.~\ref{table:ablation-loss}, our SHM can largely improves the model performance even without using the proposed dual consistency learning. This demonstrates the importance of training the model with diverse samples and the effectiveness of our SHM.

\noindent\textbf{Effectiveness of Style Consistency (SC).}
SC is the consistency constraint that leads the model to learn style invariant representation. In Tab.~\ref{table:ablation-loss}, compared with only applying SHM, SC yields an improvement of 3.99\% in average mIoU, demonstrating the superiority of the proposed logit pairing over cross entropy loss in learning style invariant model.

\noindent\textbf{Effectiveness of Retrospection Consistency (RC).}
\textit{First}, RC serves as an important guidance for narrowing the domain gap between synthetic and real data. Applying RC on top of SHM can yield an improvement of 4.61\%,
while removing RC will degrade the performance of \ours by 2.75\% in mIoU. 
\textit{Second}, we conduct experiments to verify that the effectiveness of RC lies in the real-world knowledge instead of feature-level distance minimization of the paired samples. 
As directly minimizing the feature-level absolute distance of paired samples will lead to sub-optimal results (lead all the features close to zero), we replace the ImageNet pre-trained model in RC by exponential moving average (EMA) model. Comparing the fifth row and the sixth row in Tab.~\ref{table:ablation-loss}, EMA model only gains 1.25\% improvement while RC improves the SC model by 2.75\%. The results verify the significance of the retrospective knowledge in RC.

\begin{table}[t]
\begin{center}
\caption{Comparison of different style variation methods. All models use ResNet-50 backbone and are trained with GTAV training set.}
\label{table:style-variation}
\footnotesize
\setlength{\tabcolsep}{3pt}
\begin{tabular}{l|p{2cm}<\centering|p{2cm}<\centering|p{2cm}<\centering|p{2cm}<\centering}
\toprule
Methods (GTAV) & CityScapes & BDD100K & Mapillary & Mean \\
\midrule
Baseline & 28.95 & 25.14 & 28.18 & 27.42\\
\midrule
Random Style & 37.99 & 37.63 & 38.06 &	37.89 \\
MixStyle~\cite{zhou2021mixstyle} & 43.14 & 37.94 & 42.22 & 41.10 \\
CrossNorm~\cite{crossnorm} & 43.13	&	37.20	&	41.83	& 40.72 \\
Kmeans Basis & 40.50 & 37.62 & 39.46 & 39.19 \\
\bf Ours & \textbf{44.65} & \textbf{39.28} & \textbf{43.34} & \textbf{42.42} \\
\bottomrule
\end{tabular}
\end{center}
\end{table}

\subsection{Further Evaluation}
In this section, we compare other style variation methods with SHM and evaluate two important factors influencing SHM, \ie, basis style selection interval $k$ and the insert location of SHM.

\noindent\textbf{Comparison of different style variation methods.}
We compare SHM with random style, MixStyle~\cite{zhou2021mixstyle}, CrossNorm~\cite{crossnorm} and style hallucination with Kmeans basis in Tab.~\ref{table:style-variation}. Random style utilizes the randomly sampled new styles from the standard normal distribution to form new samples.
MixStyle~\cite{zhou2021mixstyle} generates new styles by mixing the original style with the random shuffled style within a mini-batch, and CrossNorm~\cite{crossnorm} swaps the original style with another style within the shuffled mini-batch. 
SHM and Kmeans basis both use the linear combination of basis style to generate new styles, but the basis styles of SHM are selected by FPS~\cite{qi2017pointnet++} while those of Kmeans basis are obtained by Kmeans clustering centers. We can make four observations from Tab.~\ref{table:style-variation}.
\textbf{First}, we cannot make full use of dual consistency to achieve significant performance with the unrealistic random styles since standard normal distribution cannot represent the source nor the target domains.
\textbf{Second}, despite the use of realistic source styles, random utilization of MixStyle and CrossNorm leads to the generation of more samples from the dominant styles that may be different from the real-world target styles. When using MixStyle and CrossNorm, the model achieves an average mIoU of 41.10\% and 40.72\%, respectively. 
\textbf{Third}, as shown in Fig.~\ref{fig:visual-dist}(b), Kmeans basis suffers from the similar but more severe dominant style issue in style generation. As a result,
rare styles are discarded and thus the model achieves poorer performance than the above two.
\textbf{Fourth}, SHM selects basis styles with FPS, and thus the selection can cover the source distribution to  
a large extent, especially those rare styles. With such basis styles, SHM generates styles from all the source distributions, and some generated styles are even close to the target domains (Fig.~\ref{fig:visual-dist}(c)). Consequently, combining SHM with dual consistency learning, \ours can reap the benefit of the source data and outperforms other methods on all three target datasets.

\begin{figure}[t]
    \centering
    \includegraphics[width=.9\textwidth]{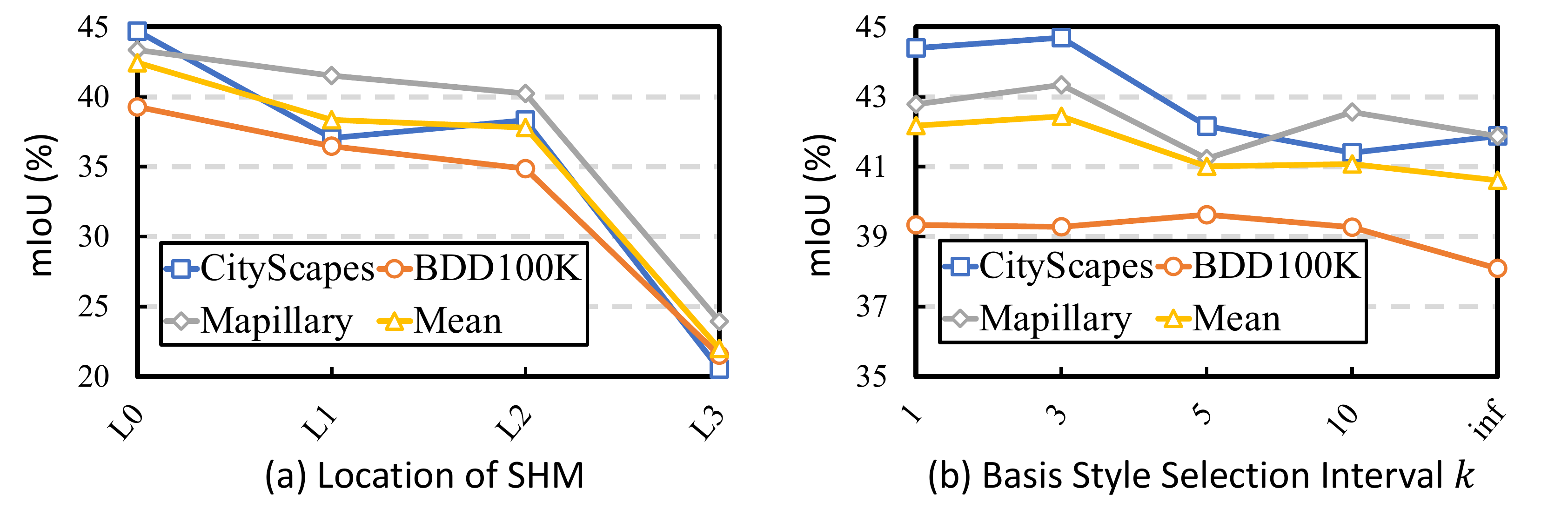}
    \caption{Parameter analysis on the location of SHM and the basis style selection interval.}
    \label{fig:param}
\end{figure}

\noindent\textbf{Location of SHM.}
We investigate the impact of inserting SHM in different locations in Fig.~\ref{fig:param}(a). ``L0'' denotes inserting SHM after the first Conv-BN-ReLU layer (layer0) and ``L1'' to ``L3'' denote inserting SHM after the corresponding (1-3) ResNet layer. As shown in Fig.~\ref{fig:param}(a), ``L0'' achieves the best result while the performance of ``L1'' and ``L2'' drops a little. However, the model suffers from drastic performance degradation when inserting SHM after layer3. The reasons are two-fold. First, the channel-wise mean and standard deviation represent more style information in the shallow layers of deep neural networks while they contain more semantic information in deep layers~\cite{adain,dumoulin2016learned}. Second, the residual connections in ResNet will lead the ResNet activations of deep layers to have large peaks and small entropy, which makes the style features biased to a few dominant patterns instead of the global style~\cite{wang2021rethinking}.

\noindent\textbf{Basis style selection interval.}
The distribution of source styles is varied along with the model training. 
To better represent the style space, we re-select the basis styles with the interval of $k$ epochs. 
The abscissa of Fig.~\ref{fig:param}(b) denotes the selection interval $k$ and ``inf'' denotes only selecting the basis style once in the beginning of training.
As shown in Fig.~\ref{fig:param}(b), the model achieves consistent and good performance with frequent re-selection ($k<=3$) while the performance degrades with the increase of selection interval, and the average mIoU is lower than 41\% when only selecting once. Taking both the performance and computational cost into consideration, we set $k=3$ in \ours.

\subsection{Real-to-Others Domain Generalization}
To further demonstrate the effectiveness of \ours, we leverage CityScapes~\cite{cityscapes} as the source domain and generalize to real (BDD100K~\cite{bdd} and Mapillary~\cite{mapillary}) and synthetic (GTAV~\cite{gtav} and SYNTHIA~\cite{synthia}) domains.
As shown in Tab.~\ref{tab:city}, \ours consistently outperforms ISW~\cite{robustnet} and IBN-Net~\cite{ibn} on both real and synthetic datasets. These results verify the versatility of our method.

\begin{table}[t]
\caption{Comparison with state-of-the-art methods trained on CityScapes.}
\footnotesize
    \centering
    \begin{tabular}{l|cccc}
    \toprule
      Methods & GTAV & SYNTHIA & BDD100K & Mapillary \\
    \midrule
    IBN-Net~\cite{ibn} & 45.06 & 26.14 & 48.56 & 57.04 \\
    ISW~\cite{robustnet}  & 45.00 & 26.20 & 50.73 & 58.64 \\
    \textbf{Ours}  & \textbf{48.61} & \textbf{27.62} & \textbf{50.95} & \textbf{60.67} \\
    \bottomrule
    \end{tabular}
    \label{tab:city}
\end{table}

\section{Conclusion}
In this paper, we propose a novel framework~(\ours) for synthetic-to-real domain generalized semantic segmentation. \ours leverages two consistency constraints to learn the domain-invariant representation by seeking consistent representation across styles and the guidance of retrospective knowledge. In addition, the style hallucination module (SHM)  is equipped into our framework, which can effectively catalyze the dual consistency learning by generating diverse and realistic source samples. Experiments on three real-world dataset show that \ours achieves state-of-the-art performance under both single- and multi-source domain generalization settings with different backbones.

\noindent\textbf{Acknowledgements.}
This research/project is supported by the National Research Foundation Singapore and DSO National Laboratories under the AI Singapore Programme (AISG Award No: AISG2-RP-2020-016), the Tier 2 grant MOE-T2EP20120-0011 from the Singapore Ministry of Education, and the EU H2020 project AI4Media (No. 951911). 

\clearpage
%
%
\clearpage

\bibliographystyle{splncs04}
\bibliography{references}

\clearpage

\appendix

\section{Visualization}
\label{sec:visualization}

\begin{figure}[ht]
    \centering
    \includegraphics[width=.99\textwidth]{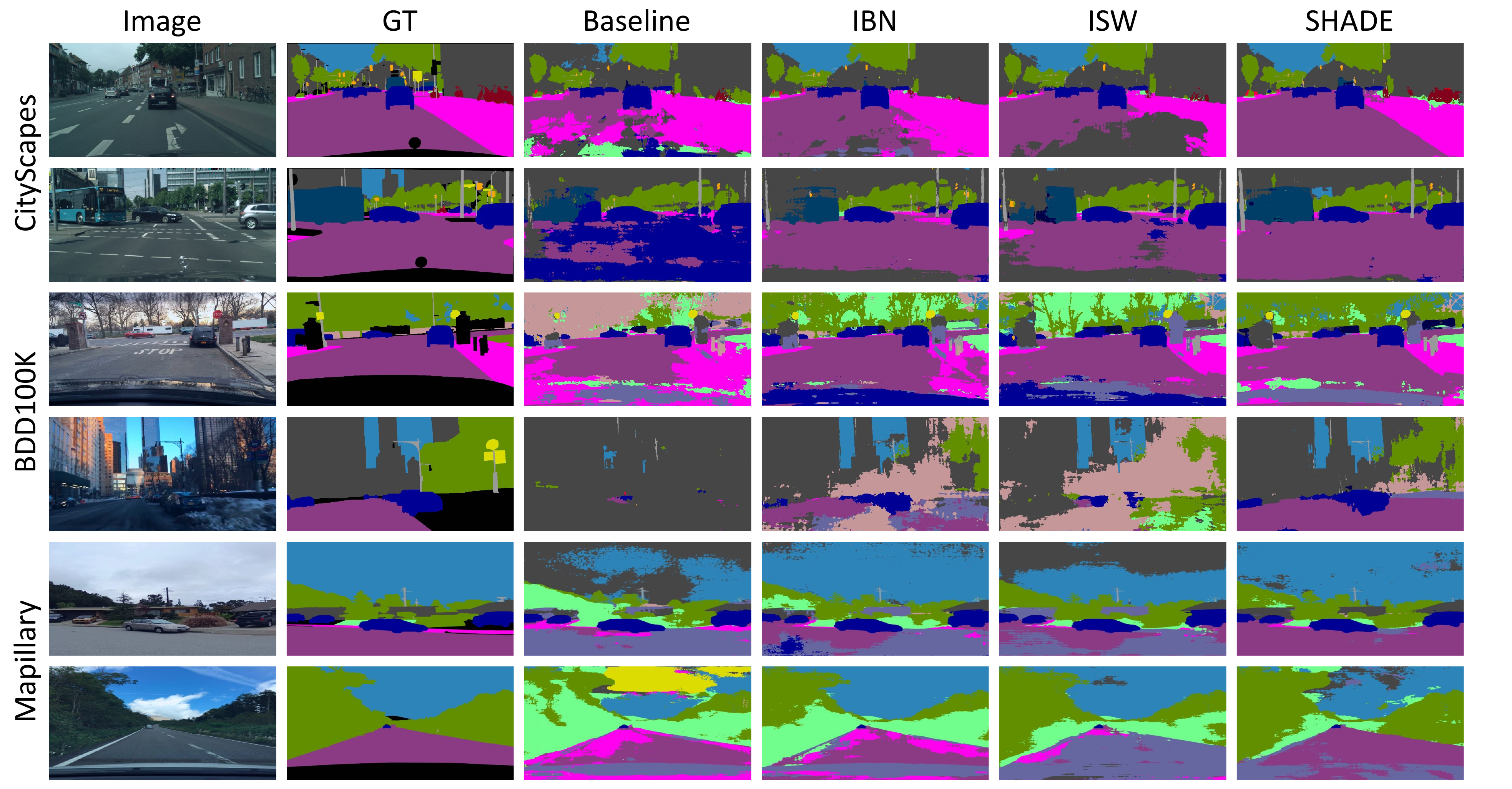}
    \caption{Qualitative comparison of segmentation results.}
    \label{fig:seg_results}
\end{figure}

\noindent\textbf{Qualitative results.} We compare the segmentation results among baseline, IBN-Net~\cite{ibn}, ISW~\cite{robustnet} and \ours on CityScapes~\cite{cityscapes}, BDD100K~\cite{bdd} and Mapillary~\cite{mapillary} in Fig.~\ref{fig:seg_results}. 
We obtain two observations from Fig.~\ref{fig:seg_results}.
First, \ours consistently outperforms other methods under different target conditions (\eg, sunny, cloudy and overcast).
Second, \ours can well deal with both ``stuff'' classes (\eg, road) and ``things'' classes (\eg, bus and bicycle).
The above two observations demonstrate that \ours is robust to style variation and has strong ability in segmenting unseen real-world images.

\begin{figure}[t]
    \centering
    \includegraphics[width=.99\textwidth]{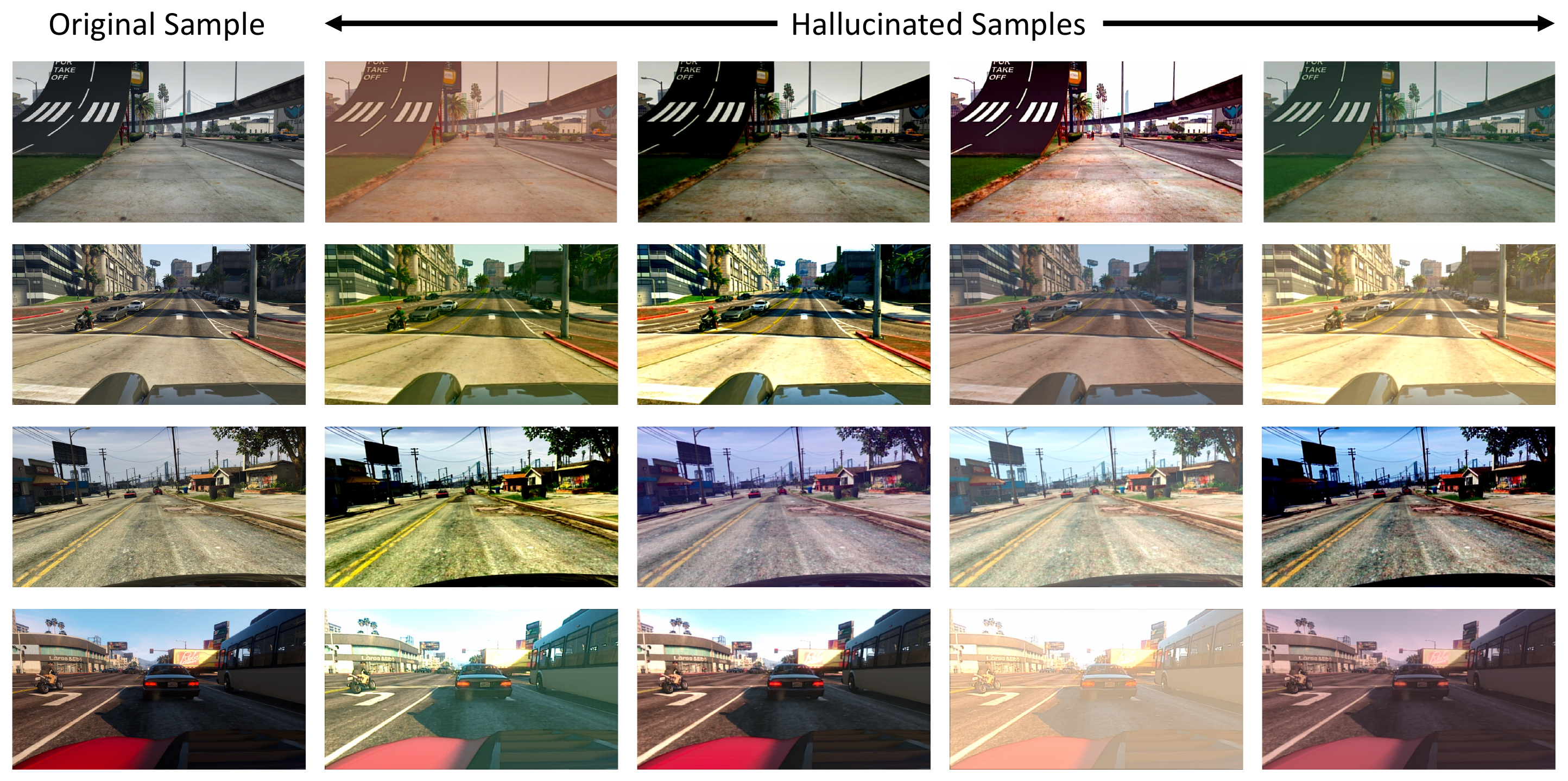}
    \caption{Visualization of style-diversified samples.}
    \label{fig:visual_style}
\end{figure}

\noindent\textbf{Style Visualization.} To better understand our SHM, we visualize the style-diversified samples with an auto-encoder.
Examples are shown in Fig.~\ref{fig:visual_style}. SHM replaces the original style features with the combination of basis styles to obtain new samples of different styles, \eg, weather change (from overcast to sunny) and time change (from dusk to midday).

\section{Per-class IoU}
We show the per-class IoU in Tab.~\ref{tab:per-class} to verify the effectiveness of retrospection consistency (RC). RC can improve the performance of both ``things'' classes (\textbf{+2.47\%}) and ``stuff'' classes (\textbf{+1.54\%}), and the improvement on the ``things'' classes is more significant.

\begin{table*}[ht]
\centering
\caption{Per-class IoU on CityScapes dataset. Models are trained on GTAV with ResNet-50 backbone.}
\label{tab:per-class}
\setlength{\tabcolsep}{3pt}
\resizebox{\textwidth}{!}{%
\begin{tabular}{l|ccccccccccccccccccc|c}
\toprule
 & \rotatebox{90}{Road} & \rotatebox{90}{S.walk} & \rotatebox{90}{Build.} & \rotatebox{90}{Wall} & \rotatebox{90}{Fence} & \rotatebox{90}{Pole} & \rotatebox{90}{Tr.Light} & \rotatebox{90}{Sign} & \rotatebox{90}{Veget.} & \rotatebox{90}{Terrain} & \rotatebox{90}{Sky} & \rotatebox{90}{Person} & \rotatebox{90}{Rider} & \rotatebox{90}{Car} & \rotatebox{90}{Truck} & \rotatebox{90}{Bus} & \rotatebox{90}{Train} & \rotatebox{90}{M.bike} & \rotatebox{90}{Bike} & mIoU\\
\midrule
\ours w.o. RC & 81.4 & 36.3 & 75.1 & \textbf{27.2} & 25.5 & 33.0 & \textbf{34.8} & \textbf{19.1} & 84.9 & 35.5 & 65.3 & 64.7 & 26.5 & \textbf{85.2} & 23.7 & 28.7 & 5.8 & \textbf{32.6} & 25.2 & 42.7 \\
\ours & \textbf{82.8} & \textbf{37.6} & \textbf{77.1} & {26.0} & \textbf{29.3} & \textbf{33.3} & {33.5} & {17.9} & \textbf{85.6} & \textbf{36.1} & \textbf{69.4} & \textbf{67.0} & \textbf{30.8} & {84.8} & \textbf{31.9} & \textbf{31.4} & \textbf{15.5} & {31.2} & \textbf{27.0} & \textbf{44.6}\\
\bottomrule
\end{tabular}
}
\end{table*}

\end{document}